\documentclass{bmvc2k}

\title{Key Person Aided Re-identification in \\
Partially Ordered Pedestrian Set}

\addauthor{Chen Chen}{chen.chen@ia.ac.cn}{1}
\addauthor{Min Cao}{caomin2014@ia.ac.cn}{1}
\addauthor{Xiyuan Hu}{xiyuan.hu@ia.ac.cn}{1}
\addauthor{Silong Peng}{silong.peng@ia.ac.cn}{12}

\addinstitution{
 Institute of Automation\\
 Chinese Academy of Sciences\\
 Beijing, China
}
\addinstitution{
 Beijing Visytem Co. Ltd.\\
}

\runninghead{ Chen \lowercase{\etal}}{Key Person Aided Re-id in Partially Ordered Pedestrian Set}

\def\eg{\emph{e.g}\bmvaOneDot}

\def\etal{\emph{et al}\bmvaOneDot}

\usepackage{outlines}
\usepackage{times}
\usepackage{epsfig}
\usepackage{graphicx}
\usepackage{amsmath}
\usepackage{amssymb}
\usepackage{multirow}
\usepackage{makecell}
\usepackage{subfigure}
\usepackage{float}
\usepackage{bm}
\usepackage{tabularx}
\usepackage{array}

\DeclareMathOperator*{\argmin}{arg\,min} 

\begin{document}

\maketitle

\begin{abstract}
Ideally person re-identification seeks for perfect feature representation and metric model that re-identify all various pedestrians well in non-overlapping views at different locations with different camera configurations, which is very challenging. However, in most pedestrian sets, there always are some outstanding persons who are relatively easy to re-identify. 
Inspired by the existence of such data division, we propose a novel key person aided person re-identification framework based on the re-defined partially ordered pedestrian sets. 
The outstanding persons, namely ``key persons'', are selected by the $K$-nearest neighbor based saliency measurement. The partial order defined by pedestrian entering time in surveillance associates the key persons with the query person temporally and helps to locate the possible candidates.
Experiments conducted on two video datasets show that the proposed key person aided framework outperforms the state-of-the-art methods and improves the matching accuracy greatly at all ranks. 
% Ideally person re-identification seeks for perfect feature representation and metric model that re-identify all various pedestrians identically in views with
% different camera configurations at different locations, which is very challenging. However, in most pedestrian sets, there always are some salient persons who are relatively easy to re-identify and some who are not. It is intuitive to re-identify these salient persons first and utilize them to help re-identifying other pedestrians. In this paper we propose a novel key person aided person re-identification framework based on the re-defined partially ordered pedestrian sets. 
% The key persons are selected by the $K$-nearest neighbor based saliency score, and the partial order in pedestrian set associates key persons and query person temporally and help to locate the possible candidates.
% Experiments conducted on two video datasets show that the our key person based framework outperforms the state-of-the-art methods and improve the matching accuracy greatly at all ranks. 

\end{abstract}

%-------------------------------------------------------------------------
\section{Introduction}
\label{sec:intro}

   Person re-identification (re-id) aims to match pedestrians across non-overlapping camera views. 
   Due to the variation in viewpoints, illumination, background and occlusions, the appearances of the same person observed in different camera views are often ambiguous and unreliable for being re-identified. 
   Ideally person re-id research seeks for perfect feature representation and metric model that can re-identify all various query persons in views of different camera configurations from different locations, which is very challenging. 
   
   However, we notice that in almost all person re-id datasets, there always are some pedestrians who are relatively easy to be re-identified accurately and some who are not. 
   Persons who are ``outstanding'' are usually re-identified with high rankings, on the contrary pedestrians who are not unique are usually wrongly matched even using powerful algorithms.
   Although pedestrians are different from dataset to dataset, there always exists such devision.

   One intuitive idea is first re-identifying the ``outstanding'' persons; then utilizing those re-identified ``outstanding'' pedestrians to help re-identify other pedestrians. 
   To realize this idea we need to solve two problems: how to define and select these ``outstanding'' persons and given the ``outstanding'' persons and how to associate and re-identify other pedestrians. 

   The persons that we attempt to find are ``outstanding'', ``unique'', easily recognized and accurately re-identified, which we call ``\textit{key persons}'' in this paper. In the sense of vision saliency, the key persons are salient persons among the pedestrian set. The $K$-nearest neighbor distance is used to measure and select key persons in this paper. 
   In order to associate key persons with other pedestrians, we use the entering time of pedestrian in camera view to define a temporal partial order. The video clip is synopsized into a pedestrian flow and the pedestrians in flow are associated with each other temporally. The temporal distance between them that measured by the difference of their entering time is used to fine locate the possible candidates.

\begin{figure}[t]
         \centering
         \subfigure[The traditional way of representing the probe/gallery set and addressing person re-identification problem;]{
         \includegraphics[width =\linewidth, height=0.2\linewidth]{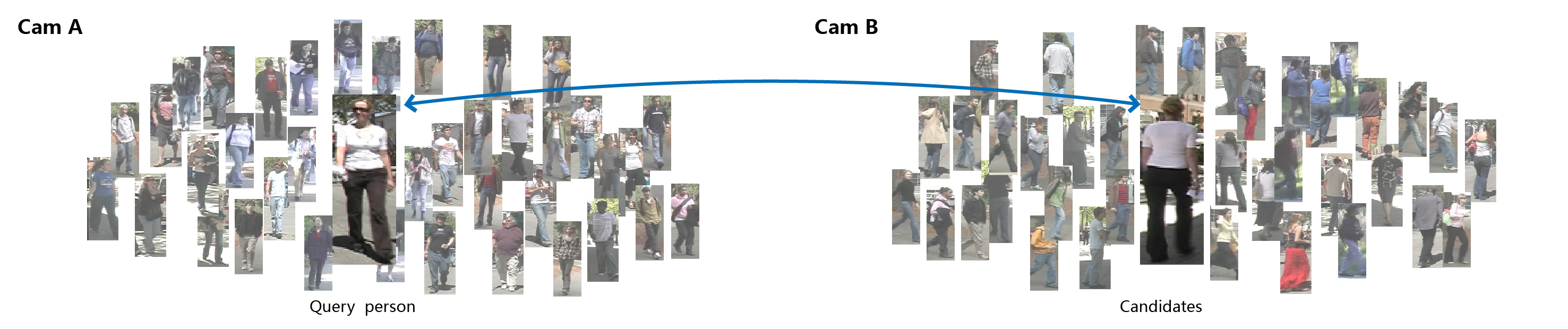}
         }%

         \subfigure[re-defining the pedestrian set as partially ordered set modeled by the temporal relations among pedestrian.]{
         \includegraphics[width =\linewidth, height=0.21\linewidth]{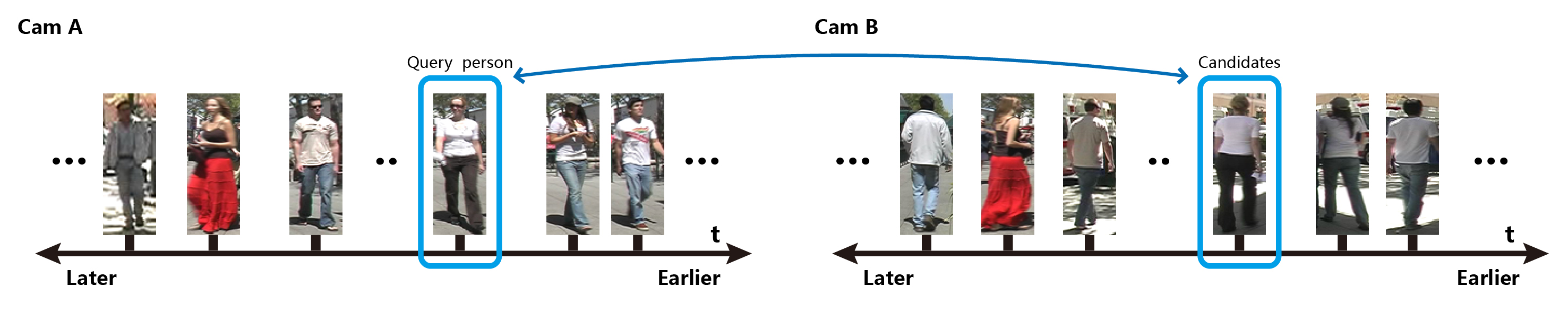}
         }
         \caption{From the traditional pedestrian set to temporally partially ordered set}
         \label{fig:intro1}
         \end{figure}

   As illustrated in Fig.\ref{fig:intro1}, the traditional way of representing pedestrian set is re-defined as a pedestrian flow with a temporal partial order, person re-id problem is addressed as element matching among two partially ordered sets using their temporal relations. 
   For instance, the query person who wears a white T-shirt and dark trousers looks similar with many other pedestrians, directly re-identifying her may lead to many wrong matches. Using the proposed idea, we can select the pedestrian with red dress as a key person, re-identify her first and utilize the temporal distance between her and the query person to fine locate possible candidates in gallery set. Re-identifying aided by key persons could reduce many false matches. 

   In this paper, we introduce the definition of key persons, model the pedestrian set as a partially ordered set with temporal relations and propose a key person aided person re-id framework, which utilized the salient persons to help re-identifying other pedestrians. 
   The following of the paper is organized as follows: section \ref{sec:relatedwork} reviews the related works, section \ref{sec:keyperson} presents how to define and select key persons, section \ref{sec:method} describes the proposed key person based pedestrian re-id framework, section \ref{sec:experiment} shows the conducted experiments and the results and section \ref{sec:conclusion} concludes the paper.

         % one figure with sub figures in one row, or multi-rows
   %-------------------------------------------------------------------------
\section{Related work}
\label{sec:relatedwork}

 % In recent years, many researchers have paid close attention to person re-id and proposed different approaches to handle the challenge of person re-id caused by different poses, viewpoints, occlusions and illumination changes. In this section, we introduce a taxonomy of image-based and video-based approaches and give a brief introduction. A comprehensive review on person re-id can be found in \cite{Zheng2016Person}.

\noindent\textbf{Image-based approaches:} Researchers usually focus on developing distinctive feature representations \cite{Farenzena2010Person,Matsukawa2016Hierarchical,Ma2014Gaussian,Shi2015Transferring} and seeking discriminative distance metrics \cite{Zheng2011Person,Liao2015Person,Liao2016Efficient,Paisitkriangkrai2015Learning} based on a single individual image. For appearance modelling, Farenzena \etal \cite{Farenzena2010Person} proposed to model the human appearance based on different body parts, which are obtained by adopting perceptual principles of symmetry and asymmetry. Matsukawa \etal \cite{Matsukawa2016Hierarchical} employed both the mean and the covariance information of pixel features via hierarchical Gaussian distribution to describe a local region in an image. For metric learning, Zheng \etal \cite{Zheng2011Person} introduced a Probabilistic Relative Distance Comparison (PRDC) model to maximize the probability of a pair of true association having a smaller distance than that of a wrong association. A low dimensional subspace is firstly learned by cross-view quadratic discriminant analysis and simultaneously a QDA metric is performed on the subspace in \cite{Liao2015Person}. Recently, deep Learning, as a powerful tool in computer vision, has attracted increasing attention for person re-id \cite{Li2014DeepReID,cheng2016person,Shi2016Embedding,Varior2016Gated} and achieved better performance. 

\noindent\textbf{video-based approaches:} Multiple images of the same person from video have been utilized for person re-identification \cite{Bk2012Boosted,Bazzani2012Multiple,Zhang2017Video,Gheissari2006Person,Ukita2016People}. Compared with a single individual image, multiple images of a person provide more clues to differentiate pedestrians from each other. For example, the appearance features extracted by multiple images were accumulated or averaged into a single signature \cite{Bk2012Boosted,Bazzani2012Multiple}. The temporal correlation of multiple images was also exploited to build the spatial-temporal appearance representation for person re-id \cite{Zhang2017Video,Gheissari2006Person}. 

In addition, the group context of a person can provide visual clues for person re-id, and there is growing interest in utilizing group information to improve re-identification performance \cite{Cai2010Matching,Li2015Subject,Zheng2009Associating, Ukita2016People,Bialkowski2013Person}. 
Group is usually defined by sociological theory in behavior and interaction analysis, and group based methods requires three steps: group detection, group feature extraction and/or group matching. More specifically, Zheng \etal \cite{Zheng2009Associating} defined two group features Center Rectangular Ring Ratio-Occurrence Descriptor (CRRRO) which is invariant to the position changes of members in a group and Block based Ratio-Occurrence Descriptor (BRO) addressing variations in illumination and viewpoint on manually segmented group images. Cai \etal \cite{Cai2010Matching} represented the group images by the covariance descriptor, capturing the appearance and statistical properties of the group image. Li \etal \cite{Li2015Subject} computed the grouping possibility by velocity and distance between people and then detected groups by Affinity Propagation (AP) clustering algorithm. For group features they extracted both geometry and visual information of a subject's partners. These methods have shown decent results in terms of accuracies by utilizing context information of other people,  however, they work only when groups exist. The proposed key person aided framework is more flexible and practical with or without groups, and the core idea is to efficiently locate possible candidates by key persons.

%-------------------------------------------------------------------------
%-------------------------------------------------------------------------
\section{Key person: definition and selection}
\label{sec:keyperson}
 
The \textit{K} nearest neighbor ($K$-NN) distance has been used for outlier detection, clutter removal and patch saliency learning \cite{Dubuisson2002A}. The average \textit{k}-NN distance can measure how distinct the query point is from the rest points in the set, which represents a saliency measurement for point set. We utilize the $K$-NN based saliency measurement to define the key persons in a pedestrian set. In order to select sufficient number of reliable key persons, we introduce a feature bank based key person selection strategy. 

\subsection{Key person definition}
\label{subsec:keyperson_def}

   Let $\mathcal{P} $ be a pedestrian set of $N$ persons and $f$ be the feature extraction strategy that maps the person image into feature space, the similarity score between person $p_i$ and $p_j$ in feature space of $f$ using the employed metric is defined as $d_f(p_i,p_j)$. The \textit{saliency score} of person $p$ in set $\mathcal{P}$ is defined as the normalized averaged $K$-NN distance
   \begin{equation}
     s_f(p) =  \mathcal{N} \big(\frac{1}{K}\sum_{k=1}^{K}d_f(p,p_{n_k}) \big),
   \end{equation}
   % \begin{equation}\label{}
   % % s^{\text{salience}}_i = \frac{1}{k}\sum_{l=1}^{k}d(p_i,p_{n_l})
   % s_f = \widetilde{s}_f(p),
   % \end{equation}
   % where
   % \begin{equation}
   %   s_f(p) =  \frac{1}{K}\sum_{k=1}^{K}d_f(p,p_{n_k}) 
   % \end{equation}
   where $\{p_{n_k}\}_{k = 1,...,K}$ denotes the $K$-NNs of person $p\in \mathcal{P}$ and $\mathcal{N}$ denotes a min-max normalization operator to linearly scale the averaged $K$-NN distance into the range [0, 1]. Person $p$ is defined as a \textit{key person} in set $\mathcal{P}$ if the saliency score is larger than a threshold $s_f(p) \geq \rho$, and the \textit{key person set} defined in feature space of $f$ is written as $\mathcal{S} = \{p \in \mathcal{P}| s_f(p) \ge \rho\}$. The saliency threshold $\rho$ should be a relatively large value to ensure the reliablity of selected key persons. 

   We can infer from the definitions that the same person may have different saliency scores and be categorized differently as a key person or not, if employing different feature extraction strategy $f$ to represent pedestrians and using different metric scheme $d_f(\cdot,\cdot)$ to measure the similarity between features. 

\subsection{Feature bank based key person selection}
\label{subsec:keyperson_selection}

   Researchers have proposed various efficient feature representations and metrics re-identifying pedestrians descriptively and discriminatively across views. Based on a feature bank $\mathcal{F} = \{f^m, m = 1,...M\}$, $M$ key person sets $\mathcal{S}^m = \{p \in \mathcal{P}|s_{f^m}(p) > \rho^m \}$ are obtained in the feature space projected by feature $f^m$. 
   And for the pedestrian set, the whole key person set is the union of every key person set $\mathcal{S} = \cup_{m=1}^M \mathcal{S}^m$ with threshold $\rho = \{\rho_1, \rho_2,...\rho_m\}$.
   These $M$ key person sets may (partly) overlap, since some key persons may be salient in multiple feature space. The non-overlapping part suggests the whole key person set $\mathcal{S}$ is selected by complementary features in feature bank, which shows the effectiveness and necessity of introducing feature bank based key person selection.

%------------------------------------------------------------------------
\section{Key person aided pedestrian re-identification}
\label{sec:method}

   In the traditional way, the probe (or gallery) set represents only a number of pedestrian images within bounding boxes, and person re-id means matching an image in probe set with the whole set of images in gallery. However, with the temporal information from videos, the probe (or gallery) set can be defined as a temporal sequence with a partial order, and the key persons can help to locate possible candidates and reduces false matches. 

\subsection{From traditional pedestrian set to partially ordered set}
\label{subsec:method_abstraction}
   
   Let $p$ denote a person in probe set $P$ and $T$ denote the starting frame of person $p$ appeared in a camera view, a partial order for the probe set is defined by the relation of the starting frame $T$ of pedestrian, let 
   $p_i \leq p_j \text{  if  } T_i \leq T_j$. Then the probe set $P$ from a camera view is defined as a partially ordered set
   % $\mathcal{P} = \{p| p \leq p_i \text{  if  } T \leq T_i, \:\:p \in P, p_i \in P \}.$
   \begin{equation}
      \mathcal{P} = \{p| p \leq p_i \text{  if  } T \leq T_i, \:\:p \in P, p_i \in P \}.
   \end{equation}
   The temporal distance between elements $p_i$ and $p_j$ in $\mathcal{P}$ is measured by $\Delta T = T_i-T_j$. 
   Similarly, the traditional gallery set $G$ is re-defined as a partially ordered set $\mathcal{G}$. 
   In a vivid description, the pedestrian set is now re-defined as a temporally ordered pedestrian flow, and the relative temporal distance between each other is defined by the temporal partial order. 

   People appeared in the same scene usually walk with similar velocity. Under such assumption, the partially ordered pedestrian set keep the internal temporal relation; 
   when the flow passes by two adjacent cameras views, most pedestrians in the flow keep their relative partial orders. 
   The stability of relative order usually ensures the resistance of the temporal relations across adjacent camera views. 

   Consider more practical cases in real world, when pedestrian velocity vary greatly and their relative temporal partial order is hardly kept, the probe set can split into subsets due to velocity contains as $\mathcal{P} = \cup_i \mathcal{P}_{\bm{v_i}}$, and 
   \begin{equation}\label{equ:subset}
   \mathcal{P}_{\bm{v_i}} = \big\{p \in \mathcal{P} \big|  \left\| \bm{v}  \right\| \in \big[\left\| \bm{v_i} \right\|- \epsilon, \left\| \bm{v_i} \right\|+ \epsilon\big], \emph{arccos}\ (\dfrac{{\bm{v} \cdot \bm{v_i}}}{{\left\| {\bm{v}} \right\|\left\| {\bm{v_i}} \right\|}}) < \theta  \big\},
   \end{equation}
    where $\bm{v_i}$ denotes the main velocities and $\mathcal{P}_{\bm{v_i}}$ denotes the subset of pedestrians with similary velocity both in speed magnitude and walking direction. Then within each subset $\mathcal{P}_{\bm{v_i}}$ the proposed key person based person re-id framework can still apply. 
 
         \begin{figure*}
         \begin{center}
         \includegraphics[width =\linewidth, height=0.27\linewidth]{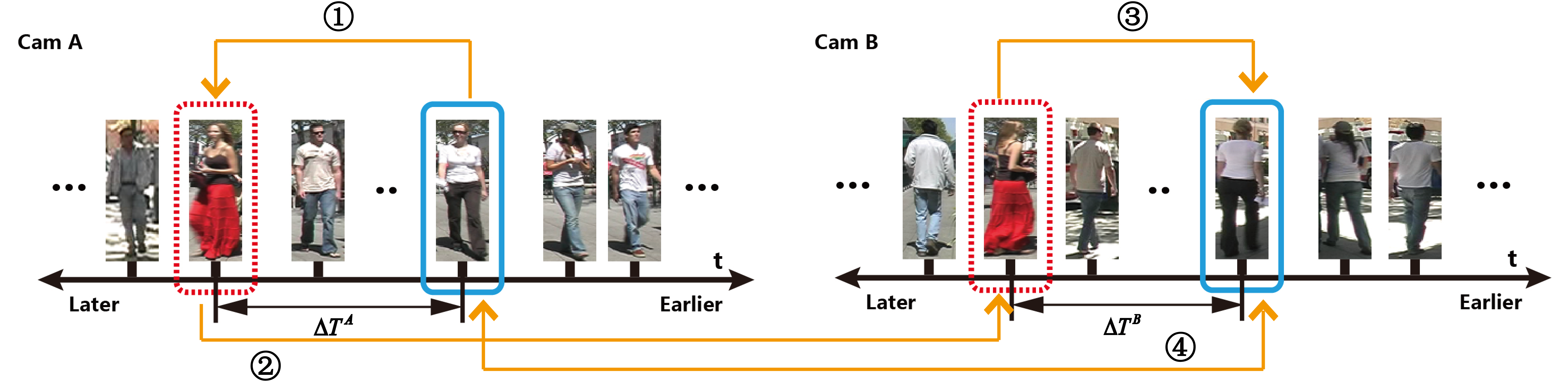}
         \end{center}
            \caption{Overview of the proposed key person aided person re-id framework with four steps: 1) find the nearest key person of query person; 2) re-identify key person with the top match in gallery set; 3) use temporal constrain to locate the possible candidate of query person; 4) weight the possible candidate and rank candidates in gallery set. Better viewed in color.}
         \label{fig-method-1}
         \end{figure*}

\subsection{Key person based person re-identification framework}
\label{subsec:setmatching}
   
   The framework of key person based person re-id consists four steps: 1) in probe set, find the nearest key person of the query person by temporal distance; 2) Match that key person with candidates in gallery set and obtain the top match; 3) Using temporal constrain provided by key person's top match to locate the possible candidates of query person; and 4) assign different weights to the possible candidates and rest pedestrians in gallery set and rank them.

      \noindent 1) $\bm{p^A} \to \bm{p}^{\bm{keyA}}$:
         Given the probe set $\mathcal{P}$ and the gallery set $\mathcal{G}$ with a partial order defined by the starting frame of pedestrian appeared in video, select the key person set $\mathcal{S} = \cup_{m=1}^M \mathcal{S}^m$ in $\mathcal{P}$ with a feature bank $\mathcal{F}=\{f^1, f^2,...f^M\}$. Let a query person be $p^A$ with the starting frame $T^A$, find the nearest key person $p^{keyA}\in \mathcal{S}$ with feature $f^m$ by the temporal distance, and compute the temporal distance between them,

            \begin{equation}
               p^{keyA} = \argmin \limits_{p_i \in \mathcal{S}} \| T^A - T^A_i\|,          
            \end{equation}

      \noindent 2) \bm{$p^{\bm{keyA}}}  \to \bm{p^{\bm{keyB}}}$: 
         Match the key person $p^{keyA}$ with the candidate $p^B \in\mathcal{G}$, compute the similarity score $d^{key}$ in feature space of $f^m$ and normalized into (0,1) as
         % $d^{key} = \widetilde{d}_{f^m}(p^{keyA},p^{keyB})$, 
         \begin{equation}
            d^{key} = \widetilde{d}_{f^m}(p^{keyA},p^B),
            \label{equ-d_key}
         \end{equation}
          and find the top match $p^{keyB}$, where $p^{keyB}$ does not denote a key person in gallery set by the proposed definition but refers a short expression for the key person $p^{keyA}$'s top match,
          	   \begin{equation}
          	   p^{keyB} = \argmin \limits_{p^B \in \mathcal{G}} \widetilde{d}_{f^m}(p^{keyA},p^B)
          	   \end{equation}

      \noindent 3) \bm{$p^{\bm{keyB}}}  \to \bm{\mathcal{G}}^*$:
         Compute the temporal distance between the key person and query person $\Delta T^A =  T^A - T^{keyA}$, and let $T^{*B}$ denote the starting frame of the query person's correspondence $p^{*B}$ in $\mathcal{G}$, and $\Delta T^{*B} =  T^{*B} - T^{keyB}$. If the temporal order of pedestrian flow across camera is strictly kept, $\Delta T^A = \Delta T^{*B}$ holds, i.e.~$T^{*B} = T^{keyB} +\Delta T^A$. In practice, the possible candidates $p^B$ should appear with large probability around the time $T^{*B}$ with a tolerance interval parameterized by $\tau$,
         % $\mathcal{G}^*= \big\{ p_j^B \in \mathcal{G} \big| T_j^B \in \big[T^B(1- \tau), T^B(1+ \tau) \big] \big\}  $.
            \begin{equation}
            \mathcal{G}^*= \big\{ p_j^{*B} \big| p_j^{*B} \in \mathcal{G},\:  T_j^{*B} \in \big[T^{keyB}+(1- \tau)\Delta T^A,\:T^{keyB}+(1+\tau)\Delta T^A \big] \big\}.   
            \end{equation}

      \noindent 4) $\bm{p^A} \to \bm{\mathcal{G}}$:
         Match the query person $p^A$ with candidate $p^B$ in gallery set $\mathcal{G}$ by assigning the original similarity scores with different weights. The original similarity score $d_{f^{base}}$ is computed using the baseline method in feature space of $f^{base}$. 
         We assign weights to the possible candidates in $\mathcal{G}^*$, compute the new similarity score for all candidates in $\mathcal{G}$ and rank them,
         
         % For the $l_{th}$ nearest key person $p^{keyA} \in \mathcal{S}^*$, we obtain a top match $p^{keyB} \in \mathcal{G}$ for $p^{keyA}$ and a possible candidate set $\mathcal{G}^*_{l}$, with the similarity score of the key person and the correspondence $\widetilde{d_l}^{key}$ computed according to Equation.\ref{equ-d_key}. 

         \begin{equation}
            d^*(p^A, p^B) = \omega \cdot d_{f^{base}}(p^A,p^B),
            \label{equ:newscore}
         \end{equation}
         where 
         \begin{equation}\label{e5}
            % \omega_i = \mathcal{N} \big(d(p^{keyA},p_i^{keyB}) \big)
            \begin{aligned}
            \omega =\left\{
            \begin{array}{ccl}
            d^{key}, & &{if\ p^B \in \mathcal{G}^*},\\
            1, & &{otherwise}.\\
            \end{array} \right.
            \end{aligned}
         \end{equation}
         denotes the weight and $d^{key}$ is defined in Eq.~(\ref{equ-d_key}). 

   In order to increase the matching accuracy for the query person, we usually take multiple key persons close to the query person to locate the possible candidate set. Let $p_l^{keyA}$ be the $l_{th}$ nearest key person defined by temporal distance, $l = 1,2,...,L$, and let $p_l^{keyB}$ be the top match of $p_l^{keyA}$ in $\mathcal{G}$ and the similarity score be $d^{key}_l$, the possible candidate set for query person is $\mathcal{G}^*_l$. As a result, the assigned weight for similarity score in Eq.~(\ref{equ:newscore}) is written as 

         % \begin{equation}
         %    d^*(p^A, p^B) = \omega_l \cdot d_{f^{base}}(p^A,p^B),
         % \end{equation}
         % where 
         \begin{equation}\label{e5}
            % \omega_i = \mathcal{N} \big(d(p^{keyA},p_i^{keyB}) \big)
            \begin{aligned}
            \omega =\left\{
            \begin{array}{ccl}
            d_l^{key}, & &{if\ p^B \in \mathcal{G}_l^*},\\
            1, & &{otherwise}.\\
            \end{array} \right.
            \end{aligned}
         \end{equation}  
   Then the rankings of all $p^B \in \mathcal{G}$ are computed based on the new similarity score $d^*(p^A, p^B)$.

         \begin{figure*}
         \begin{center}
         \includegraphics[width =0.9\linewidth]{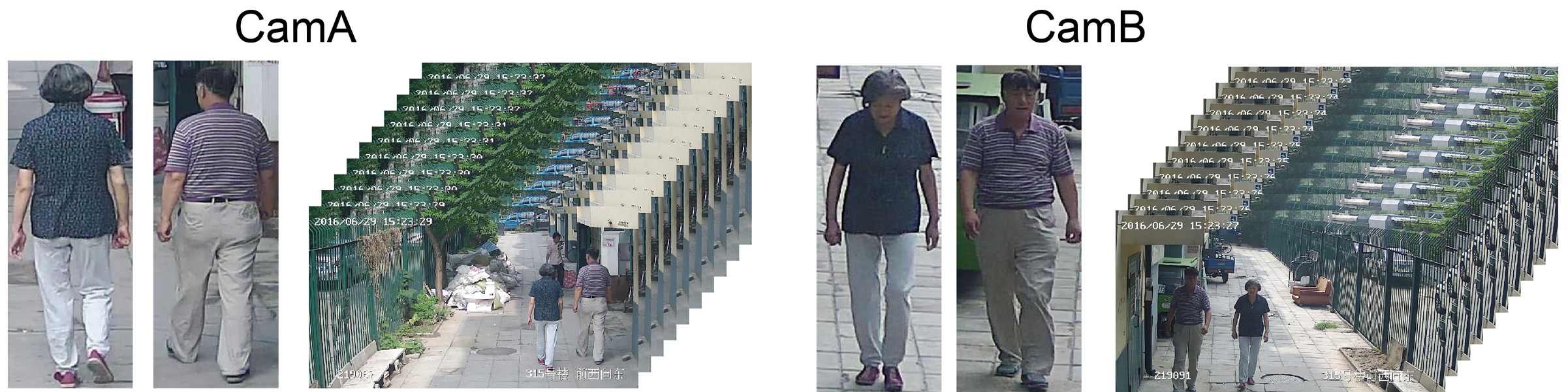}
         \end{center}
            \caption{Example of CYBJ-G dataset: two cropped person images and the corresponding video frames in different camera views.}
         \label{fig:cybjg_data}
         \end{figure*}

%------------------------------------------------------------------------
\section{Experiments and Results}
\label{sec:experiment}
   We first provide a proof of concept for key person, then evaluate the proposed framework on two datasets and show the comparisons with the state-of-the-art methods.
   Due to the limited video dataset, we evaluate our approach on a public dataset: PRID2011 dataset and a new dataset proposed by us: CYBJ-G dataset. 

   \noindent\textbf{PRID2011} dataset \cite{Hirzer2011Person} includes video and person images recorded from two cameras and full surveillance videos. 385 and 749 persons were recorded in camera views, respectively. We use the single images of the first 200 pedestrians who appear in both cameras. 

   \noindent\textbf{CYBJ-G} dataset consists 194 pedestrians captured by two surveillance cameras from the frontal and back view in a residential area. For each view, data for each pedestrian consists one cropped image and the sequential images of the corresponding video clip. The cropped images of persons have been resized to $384\times144$ pixels. Some examples are shown in Fig.~\ref{fig:cybjg_data}. 

   The feature bank employs three classic and state-of-the-art feature SDALF \cite{Farenzena2010Person}, GOG \cite{Matsukawa2016Hierarchical} and DNS \cite{Zhang2016Learning}. The baseline person re-id method adopts GOG+XQDA \cite{Matsukawa2016Hierarchical} method. 
   For all experiments, each dataset was randomly split into half for training and half for testing. All experiments were repeated for 10 trials and the results were averaged for better stability. The results are shown in standard Cumulated Matching Characteristics (CMC) curves.

    % \rho= \{\rho_{GOG},\rho_{DNS},\rho_{SDALF}\}
      % one figure with sub figures in one row, or multi-rows
      \begin{figure}[!tp]
      \centering
      \subfigure[]{
      \includegraphics[width=.35\textwidth]{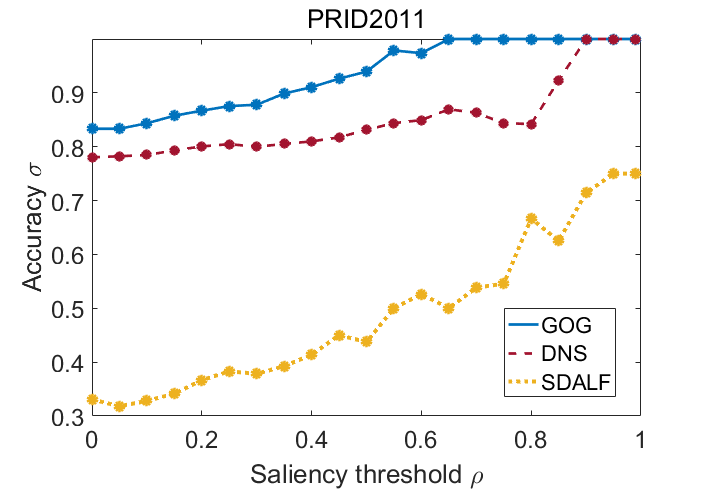}
      }%
      \subfigure[]{
      \includegraphics[width=.35\textwidth]{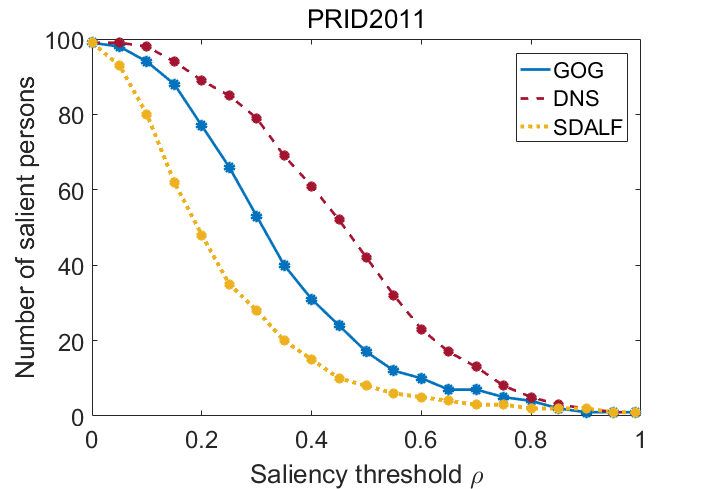}
      }%

      \subfigure[]{
      \includegraphics[width=.35\textwidth]{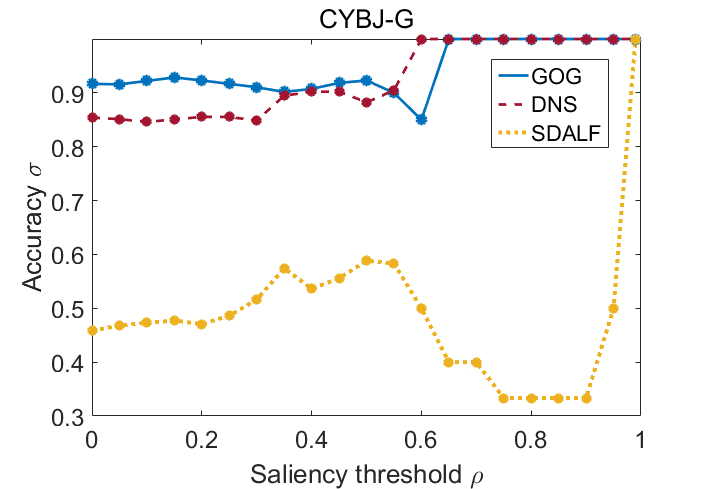}
      }%
      \subfigure[]{
      \includegraphics[width=.35\textwidth]{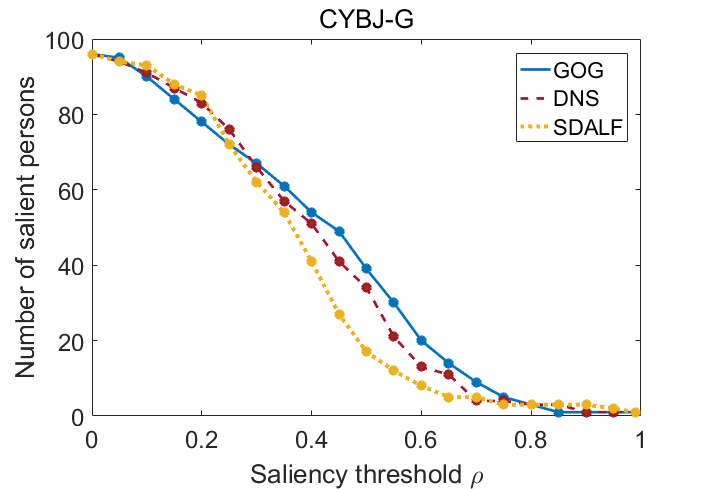}
      }
      \caption{Reference curves for saliency threshold setting: (a) the matching accuracy $\sigma$ and (b) number of key persons $N_S$ change due to the saliency threshold $\rho$ varies for PRID2011 dataset; (c-d) for CYBJ-G dataset.}
      \label{fig:rho-N}
      \end{figure}

\subsection{Proof of concept for key person} % (fold)
\label{subsec:proof}

   \noindent \textbf{Key person selection}       
      The saliency threshold $\rho$ should ensure most selected key persons are reliable matching across view. According to the key person definition, the saliency threshold $\rho$ should not be too small, otherwise many key persons that are not reliable enough will be selected; while $\rho$ should not be too large, otherwise very few key persons are selected so that the temporal distance with query person will be large and the re-id accuracy will be harmed by larger noise. 
      % Therefore, we need consider both the key person's matching accuracy and the size of key person set to set $\rho$ properly. 
      Figure.\ref{fig:rho-N} shows the matching accuracy $\sigma$ and the number of key persons $N_S$ vary due to the change of $\rho$ for three features on two datasets. As a trade off between high accuracy and reasonable size, we set the saliency thresholds $[\rho_{GOG},\rho_{DNS},\rho_{SDALF}] = [0.7,0.9,0.99]$ for PRID2011 dataset and $[0.9,0.6,0.99]$ for CYBJ-G dataset.

\begin{figure} 
            \centering \footnotesize
            \begin{tabular}{ccc ccc ccc}
            %\bmvaHangBox{\fbox{\parbox{2.7cm}{~\\[2.8mm]\rule{0pt}{1ex}\hspace{2.24mm}\includegraphics[width=2.33cm]{images/eg1_largeprint.png}\\[-0.1pt]}}}&
            \bmvaHangBox{\fbox{\includegraphics[width=1.1cm]{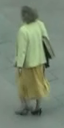} \includegraphics[width=1.1cm]{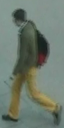} \includegraphics[width=1.1cm]{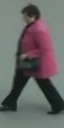}}}&

            \bmvaHangBox{\fbox{\includegraphics[width=1.1cm]{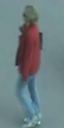} \includegraphics[width=1.1cm]{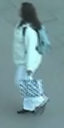} \includegraphics[width=1.1cm]{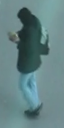}}}&

            \bmvaHangBox{\fbox{\includegraphics[width=1.1cm]{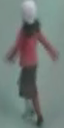} \includegraphics[width=1.1cm]{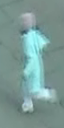} \includegraphics[width=1.1cm]{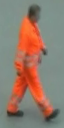}}}\\
             GOG & DNS & SDALF
            \end{tabular}
            \caption{Illustration of the key persons with saliency scores $s = 1$ in different feature space on PRID2011 dataset.}
            \label{fig:illustration1}
            \end{figure}

   \noindent \textbf{Key person illustration} 
      % Selecting the key persons by parameter setting above, we compute the saliency scores and obtain the key persons. In Fig.~\ref{fig:illustration1}, we illustrate the key persons in PRID2011 with saliency score $s=1$ using GOG, DNS and SDALF, respectively. Checking with the corresponding matches across camera views, the rankings of the true candidates are all top 1. As we can see from Fig.~\ref{fig:illustration1}, the most salient persons from different trials in PRID2011 mostly having outstanding appearance with bright color or unique accessories, which matches with the processing of human visual saliency. 
    In Fig.~\ref{fig:illustration1}, we illustrate the key persons in PRID2011 with saliency score $s=1$ using GOG, DNS and SDALF, and the rankings of their true candidates are all top 1. Most shown key persons appearance with bright color or unique accessories, which matches with the processing of human visual saliency. 
     
\begin{figure} \scriptsize \centering \renewcommand{\arraystretch}{1.5}
            \newcommand\mgape[1]{\gape{$\vcenter{\hbox{#1}}$}}
	     		\begin{tabular}{ccccccc}
      		\Xhline{1.2pt}
			    \mgape{\textbf{Person Image}} &
			    \mgape{\bmvaHangBox{\includegraphics[width=0.08\textwidth]{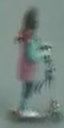}}} &
			    \mgape{\bmvaHangBox{\includegraphics[width=0.08\textwidth]{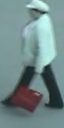}}} &
			    \mgape{\bmvaHangBox{\includegraphics[width=0.08\textwidth]{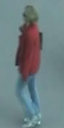}}} &
			    \mgape{\bmvaHangBox{\includegraphics[width=0.08\textwidth]{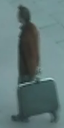}}} &
			    \mgape{\bmvaHangBox{\includegraphics[width=0.08\textwidth]{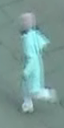}}} &
			    \mgape{\bmvaHangBox{\includegraphics[width=0.08\textwidth]{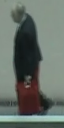}}} \\  

			    \hline
			    \textbf{Person ID} & 107 & 61 & 67 & 63 & 98 & 194 \\
			    \hline
			    \multirow{2}{*}{\textbf{Feature Score}} & 
			    \multirow{2}{*}{GOG 0.8388} &
			    \multirow{2}{*}{GOG 0.7047} &
			    GOG 0.7987 &
			    \multirow{2}{*}{GOG 0.7724} &
			    \multirow{2}{*}{SDALF 1.0} &
			    \multirow{2}{*}{GOG 1.0} \\
			     & & & DNS 1.0 & & &  \\
			    \hline
			    \textbf{ Re-ID Rank } & 1 & 2 & 1 & 1 & 1 & 1 \\
			    \Xhline{1.2pt}
			    \\
			\end{tabular}
			\caption{The illustration of key persons selected in one trial on PRID2011 dataset.}
			\label{fig:illustration2}
			\end{figure}

     Fig.~\ref{fig:illustration2} illustrate the key persons selected from a trial from experiments on PRID2011 with their feature labels, saliency scores and the rankings of their true matches in gallery set. 
      We can see that a) sometimes some key person is salient in multiple feature space and the key person sets overlap, \eg person No.~67 is selected with both GOG and DNS feature; b) the key persons selected in different feature space are mostly different, which verifies the complementary of the feature space; c) the key persons are reliable and their rankings of true matches are all top 1 except one key person. The high ranking rate ensures the correct matching of key persons as the foundation for key person based person re-id framework. 
      % In each trial with different split of dataset, the key person set may be different. We randomly choose a trial from experiments on PRID2011 and show some of the selected key persons with their feature labels, saliency scores and the rankings of their true matches in gallery set in Fig.~\ref{fig:illustration2}.
      % We can see that a) sometimes some key person is salient in multiple feature sapce and the key person sets overlap, e.g.~ person No.~67 is selected with both GOG and DNS feature; b) the key persons selected using different features are mostly different, which shows the complementary of the feature sapce; c) the key persons are reliable and their rankings of true matches are all top 1 except one key person. The high ranking rate can ensure the correct matching of key persons, which is the foundation for the performance of key person based person re-id framework. 

%---------------------------------------------------------------
\subsection{Comparison with state-of-art methods} % (fold)
\label{subsec:comparison}

   The statistics of pedestrian velocity shows that walking speeds in both datasets are in a reasonable interval, and walking directions mainly pointing in opposite two directions. According to Eq.(\ref{equ:subset}), we split the probe set into two subsets due to the walking directions. To perform the proposed framework, we also need to set the time interval parameter $\tau$ and the number of nearest key persons $L$, which rely on the stability of the temporal order among pedestrians. We set $\tau= 0.3, L = 4$ for PRID2011 and $\tau= 0.1, L = 2$ for CYBJ-G since pedestrians in CYBJ-G dataset show better temporal stability than PRID2011 dataset. 

   In Table~\ref{tab:exp_all}, we reported the comparison of the proposed method with the classic and existing state-of-the-art person re-id methods including SDALF \cite{Farenzena2010Person}, Salience \cite{Zhao2013Unsupervised}, LOMO+XQDA \cite{Liao2015Person}, SCSP \cite{Chen2016Similarity}, DNS \cite{Zhang2016Learning} and GOG \cite{Matsukawa2016Hierarchical} which are based on single static image, as well as DTDL \cite{Karanam2015Person}, PaMM \cite{Cho2016Improving} and STA \cite{Liu2015A} based on videos. 

   As shown Table \ref{tab:exp_all}, the proposed method outperforms all the classic and state-of-art methods and the performance enhancement has been achieved at all ranks $r = 1,5,10,20$. For PRID2011 dataset, Rank 1 accuracy is 81.4\% with an improvement of 22.2\% compared with the baseline method GOG and 17.3\% compared with the second best result by STA; Rank 20 accuracy is 99.7\% which means the true correspondence is almost 100\% in top 20 matches. For CYBJ-G dataset, the Rank 1 accuracy is 89.4\%, which is 12.8\% greater than the baseline (and the second best) method GOG; the Rank 5, 10 and 20 accuracies are 98.8\%, 99.9\% and 100.0\% respectively, nearly 100\% accuracy for top 5 matches and truly 100\% for top 20 matches. The proposed framework improves the person re-id accuracy greatly compared with the state-of-art methods.

         \begin{table*}
         \centering
         \caption{Comparison with the classic and existing state-of-the-art methods on PRID2011 and CYBJ-G. The best and second best results (\%) are respectively shown in red and blue. Better viewed in colour.}
         \begin{tabular*}{\textwidth}{ @{\extracolsep{\fill}} cc|c|c|c|c|c|c|c}
         \\
         \Xhline{1.2pt}
          & \multicolumn{4}{c|}{PRID2011} & \multicolumn{4}{c}{CYBJ-G} \\
         \hline
          \textbf{Methods} & r=1  & r=5  & r=10 & r=20 & r=1  & r=5  & r=10 & r=20 \\
         \hline
         SDALF \cite{Farenzena2010Person}   & 6.4 & 24.3 & 32.7 & 44.8 & 32.2 & 57.2 & 69.8 & 79.9 \\

         Salience \cite{Zhao2013Unsupervised}  & 25.8 & 43.6 & 52.6 & 62.0 & 40.8 & 64.2 & 75.5 & 83.4 \\

         LOMO+XQDA \cite{Liao2015Person}       & 39.0 & 68.0 & 83.0  & 91.0 & 67.2  & 87.4  & 91.7  & 95.2 \\

         SCSP  \cite{Chen2016Similarity}    & 12.7 & 32.7 & 51.0  & 66.0 &  21.7 &  39.1 & 50.0 & 67.4 \\

         DNS  \cite{Zhang2016Learning}       & 38.6 & 66.6 & 78.0 & 91.1 & 58.2 & 84.2 & 91.5 & 94.1 \\

         GOG  \cite{Matsukawa2016Hierarchical}  & 59.2 & 79.6 & 89.7 & {\color{blue}95.6} & {\color{blue}76.6} & {\color{blue}94.2} & {\color{blue}96.9} & {\color{blue}98.4} \\
         \hline
         DTDL \cite{Karanam2015Person}      & 41.0  & 70.0 & 78.0 & 86.0 & ~~- & ~~- & ~~- & ~~- \\

         PaMM  \cite{Cho2016Improving}      & 45.0 & 72.0 & 85.0 & 92.5 & ~~- & ~~- & ~~- & ~~- \\

         STA  \cite{Liu2015A}       & {\color{blue}64.1} & {\color{blue}87.3} & {\color{blue}89.9} & 92.0 & ~~- & ~~- & ~~- & ~~- \\
         \hline
         % CTS+SDALF (Ours)     & 8.8 & 24.8 & 35.7 & 52.0 & 38.1 & 60.8 & 68.0 & 76.3 \\

         % CTS+SCSP (Ours)      & 15.3 & 40.8 & 54.6 & 71.5 & 33.8 & 54.8 & 63.3 & 74.7 \\

         % CTS+GOG+DNS (Ours)     & {\color{red}73.5} & {\color{red}92.9} & {\color{red}96.9} & {\color{red}99.4} & {\color{blue}81.7} & 91.5 & 92.4 & 95.1 \\

         % Ours      & {\color{red}75.2} & {\color{red}93.5} & {\color{red}97.1} & {\color{red}99.5} & {\color{red}90.1} & {\color{red}98.8} & {\color{red}99.4} & {\color{red}99.6} \\
         Ours      & {\color{red}81.4} & {\color{red}96.2} & {\color{red}98.7} & {\color{red}99.7} & {\color{red}89.4} & {\color{red}98.8} & {\color{red}99.9} & {\color{red}100.0} \\
         \Xhline{1.2pt}
         \end{tabular*}
         %\end{center}
         \label{tab:exp_all}
         \end{table*}

% subsubsection subsubsection_name (end)
%------------------------------------------------------------------------
\section{Conclusion}
\label{sec:conclusion}

In this paper, we proposed a novel key person based person re-id framework. It tackles the person re-id difficulty in significantly different way from the existing methods: 1) it models the pedestrian set as temporally ordered pedestrian flow, and the temporal orders between pedestrians keep relatively stable when the flow passes adjacent camera views by pedestrian velocity constrain; 2) key persons selected and first re-identified, then help to locate the possible candidates of the query person using their temporal distance defined by the difference of their entering time.  
The experiments show the proposed framework greatly improve the re-id accuracy and outperforms all the existing state-of-the-art person re-id methods. 
% In this paper, we proposed a novel key person based person re-id framework, in which we select key persons based on visual saliency, redefines the pedestrian set as a partially ordered set and utilize the temporal distance between the key person and query person to locate the possible candidates. The proposed framework tackles the person re-id difficulty significantly different from the existing methods: 1) it models the pedestrian set as temporally ordered pedestrian flow, and the temporal orders between pedestrians keep relatively stable when the flow passes from a camera to the adjacent one by pedestrian velocity constrain; 2) key persons selected and first re-identified in the flow, then help to locate the possible candidates to re-identify the query person. 

\section*{Acknowledgment}

We would like to thank Dr.~Shijun Wang and Dr. Jianbin Jia for the fruitful discussion, and Zhenfeng Fan for carefully proof reading. This work is supported by the National Key R\&D Program of China under Grant 2017YFC0803505, China Postdoctoral Science Foundation, and National Natural Science Foundation of China with Grant No.61571438.

% \begin{figure*}
% \begin{center}
% \fbox{\rule{0pt}{2in} \rule{0.9\linewidth}{images/eg1_largeprint.png}{0pt}}
% \end{center}
%    \caption{Example of a short caption, which should be centered.}
% \label{fig:short}
% \end{figure*}

% \begin{figure}
% \begin{tabular}{ccc}
% \bmvaHangBox{\fbox{\parbox{2.7cm}{~\\[2.8mm]
% \rule{0pt}{1ex}\hspace{2.24mm}\includegraphics[width=2.33cm]{images/eg1_largeprint.png}\\[-0.1pt]}}}&
% \bmvaHangBox{\fbox{\includegraphics[width=2.8cm]{images/eg1_largeprint.png}}}&
% \bmvaHangBox{\fbox{\includegraphics[width=5.6cm]{images/eg1_2up.png}}}\\
% (a)&(b)&(c)
% \end{tabular}
% \caption{It is often a good idea for the first figure to attempt to
% encapsulate the article, complementing the abstract.  This figure illustrates
% the various print and on-screen layouts for which this paper format has
% been optimized: (a) traditional BMVC print format; (b) on-screen
% single-column format, or large-print paper; (c) full-screen two column, or
% 2-up printing. }
% \label{fig:teaser}
% \end{figure}

% \begin{table}
% \begin{center}
% \begin{tabular}{|l|c|}
% \hline
% Method & Frobnability \\
% \hline\hline
% Theirs & Frumpy \\
% Yours & Frobbly \\
% Ours & Makes one's heart Frob\\
% \hline
% \end{tabular}
% \end{center}
% \caption{Results.   Ours is better.}
% \end{table}

\bibliography{egbib}
\end{document}